\documentclass{article}
\usepackage{iclr2026/iclr2026_conference,times}

\usepackage{amsmath,amsfonts,bm}
\usepackage{graphicx}
\usepackage{booktabs}
\usepackage{multirow}
\usepackage{hyperref}
\usepackage{url}

\title{One Language, Two Scripts: \\Probing Script-Invariance in LLM Concept Representations}
\iclrfinalcopy

\author{Sripad Karne \\
Columbia University}

\begin{document}

\maketitle

\begin{abstract}
Do the features learned by Sparse Autoencoders (SAEs) represent abstract meaning, or are they tied to how text is written? We investigate this question using Serbian digraphia as a controlled testbed: Serbian is written interchangeably in Latin and Cyrillic scripts with a near-perfect character mapping between them, enabling us to vary orthography while holding meaning exactly constant. Crucially, these scripts are tokenized completely differently, sharing no tokens whatsoever. Analyzing SAE feature activations across the Gemma model family (270M--27B parameters), we find that identical sentences in different Serbian scripts activate highly overlapping features, far exceeding random baselines. Strikingly, changing script causes less representational divergence than paraphrasing within the same script, suggesting SAE features prioritize meaning over orthographic form. Cross-script cross-paraphrase comparisons provide evidence against memorization, as these combinations rarely co-occur in training data yet still exhibit substantial feature overlap. This script invariance strengthens with model scale. Taken together, our findings suggest that SAE features can capture semantics at a level of abstraction above surface tokenization, and we propose Serbian digraphia as a general evaluation paradigm for probing the abstractness of learned representations.

\end{abstract}

\section{Introduction}

Large language models have transformed how people across the globe access and process information. Users from diverse linguistic backgrounds now interact with these systems daily, raising fundamental questions about how LLMs represent meaning across different languages and writing systems. Sparse Autoencoders (SAEs) offer a lens into this: by decomposing neural network activations into sparse, interpretable features \citep{bricken2023monosemanticity, cunningham2023sparse}, SAEs allow us to examine whether models encode meaning abstractly or remain tied to script-specific token patterns.

To explore this, we study Serbian, which is one of the few languages with active digraphia. This means it is written in two scripts: Latin and Cyrillic. Native speakers use both scripts interchangeably, and a deterministic mapping allows lossless conversion between them with zero semantic change. However, the two scripts are tokenized completely differently by LLMs, making Serbian an ideal testbed for our investigation. If SAE features capture abstract semantics, we would expect the same phrases in Serbian Latin and Serbian Cyrillic to activate similar features despite their divergent tokenization.

We systematically evaluate this hypothesis across the Gemma model family \citep{gemmateam2024gemma}, spanning 270M to 27B parameters, using Gemma Scope 2 SAEs \citep{gemmascope2}. Our experimental design compares SAE feature activations across carefully constructed comparison types, including within-script semantic comparisons (original vs.\ paraphrase), cross-script identity comparisons (same sentence in both scripts), and multiple random baselines.

Our contributions are:
\begin{enumerate}
    \item We introduce Serbian digraphia as a controlled evaluation paradigm for assessing whether learned concept representations capture abstract semantics or remain tied to script-specific token representations.
    \item We demonstrate that SAE features in Gemma models exhibit substantial script invariance: averaged across all models, cross-script similarity between identical sentences in Serbian Latin and Serbian Cyrillic reaches $\sim$0.58 Jaccard, with cross-script cross-paraphrase at $\sim$0.47, both significantly exceeding the cross-script random baseline of $\sim$0.28.
    \item We characterize how script invariance varies across model scale, finding that larger models maintain more consistent script-independent representations.
\end{enumerate}

Our results suggest that SAE-learned concepts do capture semantic structure that transcends surface-level tokenization, with implications for understanding how neural networks represent meaning across diverse input formats.

\section{Related Work}

\paragraph{Sparse Autoencoders for Interpretability.}
Sparse Autoencoders have emerged as a key tool for mechanistic interpretability, addressing the challenge of \textit{superposition} \citep{elhage2022toy}. \citet{bricken2023monosemanticity} demonstrated that SAEs can decompose MLP activations into interpretable, monosemantic features, while \citet{cunningham2023sparse} showed that SAE features in language models correspond to human-interpretable concepts. We leverage these SAEs to investigate whether the concepts they learn exhibit invariance to orthographic variation.

\paragraph{Cross-lingual and Multilingual Representations.}
A substantial body of work has investigated whether multilingual models develop language-agnostic representations. \citet{pires2019multilingual} found that multilingual BERT exhibits surprising cross-lingual transfer, even between languages with no lexical overlap. \citet{conneau2020unsupervised} showed that cross-lingual representations emerge at scale without explicit alignment objectives, while \citet{wu2019beto} demonstrated zero-shot cross-lingual transfer across typologically diverse languages. Work on Hindi-Urdu---languages that are linguistically similar but use different scripts (Devanagari and Nastaliq)---has shown that cross-script transfer is possible but imperfect due to vocabulary differences and the lack of a clean mapping between scripts \citep{moosa2023transliteration, xhelili2024breaking}.
 Our work sidesteps such confounds entirely: Serbian digraphia provides a deterministic mapping that allows lossless conversion, enabling us to vary script while holding semantics \textit{exactly} constant.

\section{Methodology}

\subsection{Serbian Digraphia as a Controlled Testbed}

Serbian is one of few languages with active digraphia: it is written interchangeably in Latin script and Cyrillic script. What makes Serbian uniquely suited for our investigation is that \textit{both scripts are used with near-equal frequency in everyday life}. This means that in any large training corpus, the same concepts and linguistic patterns appear in both orthographic forms. Critically, while semantics remain identical, the tokenizer produces entirely different token sequences for each script. This creates an ideal controlled experiment.
\subsection{Dataset}

We construct a dataset of 30 sentence triplets, each containing:
\begin{itemize}
    \item \textbf{Original:} A natural sentence covering diverse topics (nature, daily activities, abstract concepts)
    \item \textbf{Paraphrase:} A semantically equivalent rephrasing with different lexical choices
    \item \textbf{Random:} An unrelated sentence with no semantic connection
\end{itemize}

Each triplet exists in three language variants: English, Serbian Latin, and Serbian Cyrillic. Serbian translations were carefully generated and rigorously verified for accuracy. The complete dataset comprises 270 unique sentences (30 triplets, 3 variants, 3 languages/scripts). Using LaBSE sentence embeddings \citep{feng2022labse}, we further confirmed that cross-script pairs achieve near-ceiling semantic similarity. We also controlled for potential tokenization confounds by ensuring comparable token counts across script pairs.  Additional details on translation methodology, tokenization analysis, and phrase similarity verification are provided in Appendix A.

We evaluate across the Gemma model family \citep{gemmateam2024gemma}, spanning four orders of magnitude in scale: Gemma-3-270M, Gemma-3-1B, Gemma-3-4B, Gemma-3-12B, and Gemma-3-27B. For each model, we use Gemma Scope 2 SAEs \citep{gemmascope2}---JumpReLU sparse autoencoders with 65,536 features trained on model activations. We select 3--4 layers per model spanning early, middle, and late processing stages (e.g., layers 12, 24, 31, 41 for Gemma-3-12B).

Following established practices in the SAE interpretability literature \citep{templeton2024scaling, rajamanoharan2024jumping}, we use a width of 65k features and a medium L0 sparsity level, which provides a balance between feature granularity and reconstruction quality. The activation threshold of $\tau = 0.1$ corresponds to the JumpReLU threshold used. 

\subsection{Feature Extraction Pipeline}

Given an input sentence $s$, we extract the set of active SAE features $F(s)$ as follows:
\begin{enumerate}
    \item \textbf{Tokenization:} Convert $s$ to token sequence $\mathbf{t} = (t_1, \ldots, t_n)$ using the Gemma tokenizer.
    \item \textbf{Forward pass:} Compute hidden state $\mathbf{h}^{(l)} \in \mathbb{R}^d$ at layer $l$ for the final token position.
    \item \textbf{SAE encoding:} Obtain feature activations $\mathbf{a} = \text{SAE}_{\text{enc}}(\mathbf{h}^{(l)}) \in \mathbb{R}^{65536}$.
    \item \textbf{Thresholding:} Define active feature set $F(s) = \{i : a_i > \tau\}$ where $\tau = 0.1$ (the JumpReLU activation threshold).
\end{enumerate}

We use last-token pooling rather than mean pooling, as we found empirically that it yields more robust results. This pipeline is applied identically across all five Gemma models and all tested layers, with the same threshold $\tau = 0.1$ used throughout.

\subsection{Comparison Types}

We define 14 comparison types to systematically test our hypotheses. The key comparisons are:

\paragraph{Baseline Comparisons}
\begin{itemize}
    \item \textit{Original vs.\ Paraphrase} (English, Serbian Latin, Serbian Cyrillic): Tests whether SAE features capture semantic similarity within a single script/language.
    \item \textit{Original vs.\ Random} (same variants): Establishes baseline similarity for unrelated sentences.
\end{itemize}

\paragraph{Cross-Script Comparisons (Primary Test)}
\begin{itemize}
    \item \textit{Cross-Script Original} (Serbian Latin Original vs.\ Serbian Cyrillic Original): The core test of script invariance for identical sentences.
    \item \textit{Cross-Script Paraphrase} (Serbian Latin Paraphrase vs.\ Serbian Cyrillic Paraphrase): A robustness check ensuring script invariance holds.
    \item \textit{Cross-Script Cross-Paraphrase} (Serbian Latin Original vs.\ Serbian Cyrillic Paraphrase and vice versa): Tests combined script and lexical variation.
\end{itemize}

\paragraph{Random Baselines}
\begin{itemize}
    \item \textit{Cross-Script Random} (Latin Original vs.\ Cyrillic Random and Cyrillic Original vs.\ Latin Random): Unrelated sentences across Serbian scripts.
    \item \textit{Cross-Language Random} (Serbian vs.\ English unrelated sentences): Establishes a floor for random similarity.
\end{itemize}

\subsection{Evaluation Metric}

We measure representational similarity using Jaccard similarity over active feature sets. Given two sentences $s_1$ and $s_2$ with active feature sets $F(s_1)$ and $F(s_2)$, we compute:
\begin{equation}
    J(s_1, s_2) = \frac{|F(s_1) \cap F(s_2)|}{|F(s_1) \cup F(s_2)|}
\end{equation}
Jaccard similarity ranges from 0 (no overlap) to 1 (identical feature sets). For each comparison type, we compute Jaccard similarity for all 30 sentence pairs and report the mean across these pairs.

\section{Experiments and Results}

\subsection{Evidence for Script-Invariant Semantic Representations}

We first establish that SAE features encode semantic information by comparing original sentences to their paraphrases versus unrelated random sentences. If SAE features capture meaning, semantically related pairs should activate more similar feature sets than unrelated pairs.

This prediction holds with complete consistency. Across all model-layer combinations and all three language/script conditions (English, Serbian Latin, Serbian Cyrillic), original-paraphrase similarity exceeds original-random similarity in 100\% of cases, with paraphrase similarity averaging $\sim$0.54 compared to $\sim$0.28 for random pairs. Results are provided in Appendix B.1. 

Having established that SAE features capture semantics, we now test our central hypothesis: do identical sentences in Latin and Cyrillic scripts activate similar features despite entirely different tokenization?

Table~\ref{tab:main} presents our core finding. We compare five conditions in decreasing order of expected similarity: (1) \textit{Cross-Script Original}---the same sentence rendered in both scripts, (2) \textit{Cross-Script Paraphrase}---the same paraphrase rendered in both scripts, (3) \textit{Cross-Script Cross-Paraphrase}---original in one script versus paraphrase in the other, (4) \textit{Cross-Script Random}---unrelated sentences across Serbian scripts, and (5) \textit{Cross-Language Random}---unrelated Serbian and English sentences. If representations are script-invariant, we expect high similarity for conditions (1) and (2), moderate similarity for condition (3), and low similarity for condition (4) and condition (5).

\begin{table}[h]
\centering
\caption{Evidence for script-invariant representations. Results averaged across all models and layers.}
\vspace{2mm}
\label{tab:main}
\small
\begin{tabular}{lc}
\toprule
\textbf{Comparison Type} & \textbf{Mean Jaccard Similarity} \\
\midrule
Cross-Script Original (Latin $\leftrightarrow$ Cyrillic, same sentence) & 0.58 \\
Cross-Script Paraphrase (Latin $\leftrightarrow$ Cyrillic, same paraphrase) & 0.59 \\
Cross-Script Cross-Paraphrase (Latin/Cyrillic Orig $\leftrightarrow$ Cyrillic/Latin Para) & 0.47 \\
Cross-Script Random (Latin $\leftrightarrow$ Cyrillic, unrelated) & 0.28 \\
Cross-Language Random (Serbian $\leftrightarrow$ English, unrelated) & 0.19 \\
\bottomrule
\end{tabular}
\end{table}

The results strongly support script invariance. Identical sentences across scripts achieve a Jaccard similarity of $\sim$0.58, substantially higher than the random baseline of $\sim$0.28, with cross-script paraphrase similarity (
$\sim$0.59) confirming robustness across different phrasings.

Furthermore, cross-script random similarity ($\sim$0.28) exceeds cross-language random similarity ($\sim$0.19), indicating that the model treats Serbian Latin and Serbian Cyrillic as more similar to each other than either is to English. This result is particularly striking given that the tokenizer produces entirely disjoint token vocabularies for the two Serbian scripts, with no surface-level signal that these scripts represent the same language.

The ordering \textit{Cross-Script Original} $>$ \textit{Cross-Script Paraphrase} $>$ \textit{Cross-Script Cross-Paraphrase} $>$ \textit{Cross-Script Random} $>$ \textit{Cross-Language Random} suggests that SAE features reflect a semantic hierarchy where meaning, not orthography, is the primary driver of representational similarity.

\subsection{Effect of Model Scale}
Having established script invariance across our experiments, we now examine how this property varies with model scale. Full numerical results are provided in Appendix B.2.

Figure~\ref{fig:baseline_scale} presents within-script semantic discrimination across the Gemma model family. Larger models achieve lower random baselines while maintaining comparable paraphrase similarity, resulting in greater separation between semantically related and unrelated pairs. By 27B, all three conditions (English, Serbian Latin, Serbian Cyrillic) converge to nearly identical gaps ($\sim$0.28--0.29), suggesting that at sufficient scale, the model achieves comparable semantic discrimination regardless of language or script. The slight decrease in  paraphrase similarity with scale likely reflects increased feature granularity rather than degraded understanding: larger models may develop finer-grained features that distinguish subtle differences between paraphrases, which smaller models conflate. Notably, Serbian Latin and Serbian Cyrillic exhibit remarkably similar trajectories despite disjoint tokenizations, suggesting gains in semantic encoding are not script-specific.

\begin{figure}[h]
\centering
\includegraphics[width=\linewidth]{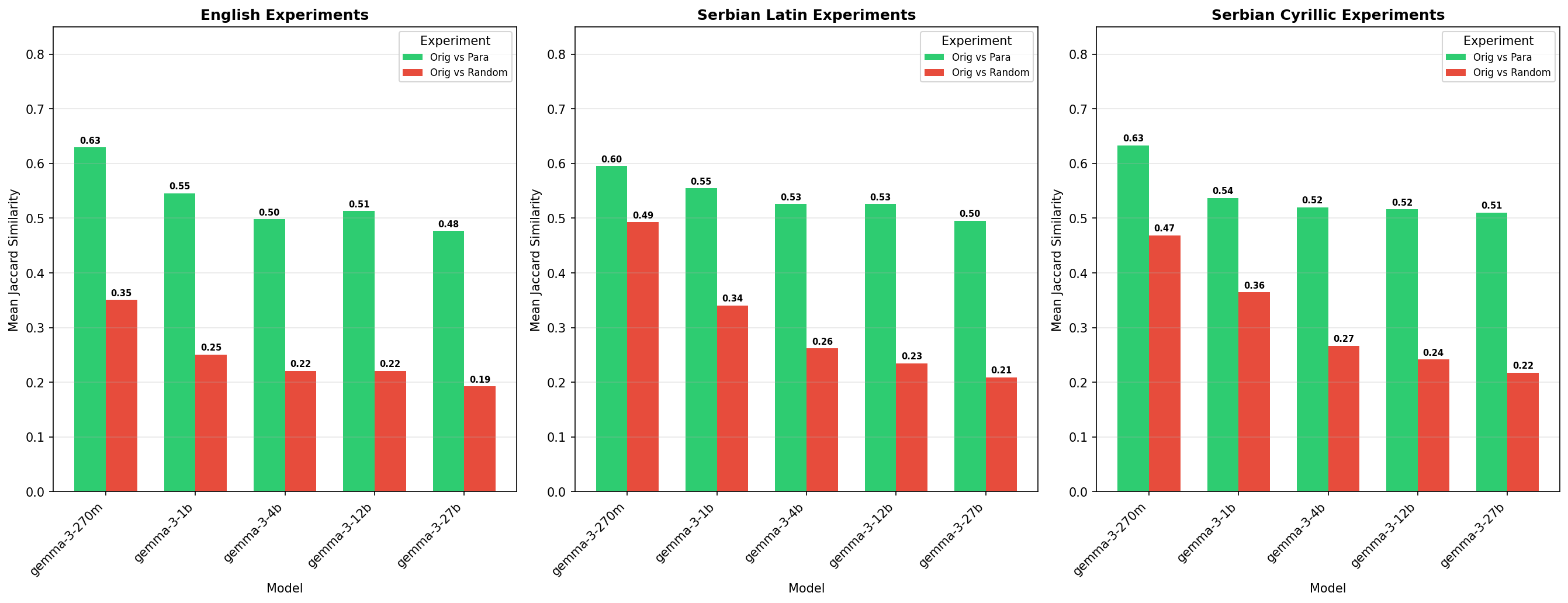}
\caption{Baseline semantic discrimination across model scales. Larger models achieve greater separation between paraphrase and random similarity, with Serbian Latin and Serbian Cyrillic following nearly parallel trajectories.}
\label{fig:baseline_scale}
\end{figure}

\begin{figure}[h]
\centering
\includegraphics[width=\linewidth]{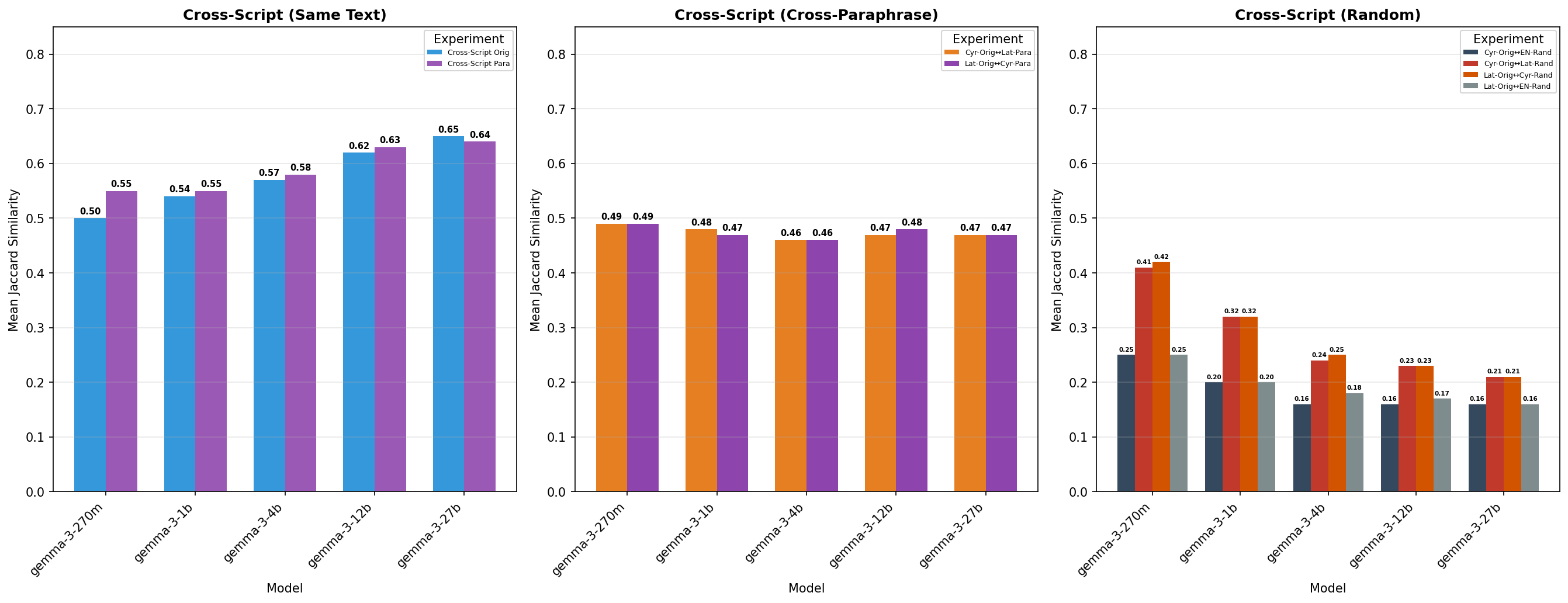}
\caption{Cross-script similarity across model scales. Cross-script original and paraphrase similarity increase with model size while cross-script random decreases, indicating increasingly robust script-invariant representations. Notably, cross-script cross-paraphrase remains stable across scales.}
\label{fig:crossscript_scale}
\end{figure}

Figure~\ref{fig:crossscript_scale} presents cross-script similarity across model scales. Cross-script original similarity (identical sentences across scripts) increases from 0.50 at 270M to 0.65 at 27B, while both random baselines move in the opposite direction: cross-script random decreases from $\sim$0.42 to $\sim$0.21, and cross-language random from $\sim$0.25 to $\sim$0.16. This pattern of semantic similarity rising while random baselines fall demonstrates that larger models develop substantially more robust script-invariant representations. Interestingly, cross-script cross-paraphrase similarity (original in one script vs.\ paraphrase in the other) remains stable across scales ($\sim$0.47--0.49). We hypothesize that two opposing effects may cancel out: larger models improve at recognizing equivalent content across scripts, but also become more sensitive to exact word choices. The net result is that cross-script cross-paraphrase similarity stays roughly constant, though further investigation is needed to confirm this explanation.

\section{Discussion}
\label{sec:discussion}

Our results demonstrate that SAE-learned features capture semantic content that transcends orthographic representation. Despite entirely disjoint tokenization, identical Serbian sentences across scripts activate highly overlapping features, and this similarity exceeds within-script paraphrase similarity, suggesting the model is more sensitive to word choice than to script. The cross-script cross-paraphrase results provide evidence against memorization-based explanations: combinations unlikely to co-occur in training data still show substantial overlap, indicating genuine semantic alignment.

These findings suggest that SAE features capture semantic content at a level of abstraction above surface orthography. If this property holds more broadly, it could have implications for cross-script interpretability research, though further work is needed to assess generalizability beyond Serbian digraphia.

\subsection{Limitations and Future Work}

Our experiments focus exclusively on Serbian digraphia. While Serbian provides an ideal controlled setting due to its deterministic script mapping and balanced real-world usage, other multi-script languages present distinct challenges. Extending this paradigm to these languages would test whether script invariance is a general property or specific to clean orthographic mappings like Serbian.

We evaluate only the Gemma model family with Gemma Scope 2 SAEs using a single configuration (65k width, medium L0, threshold 0.1). Different architectures, training procedures, or SAE hyperparameters may yield different patterns. Our dataset of 30 sentence triplets, while sufficient to establish clear statistical trends, is limited in size and domain coverage. Future work should expand to larger, more diverse corpora.

Our analysis measures feature overlap but does not establish causal relationships; future work could employ activation patching or feature ablation to verify whether shared features directly contribute to cross-script understanding. Additionally, identifying which specific SAE features are most script-invariant could reveal interpretable semantic concepts that serve as anchors for cross-lingual research. We hope Serbian digraphia, as a naturally controlled setting, proves useful for future investigations into how neural networks represent meaning across orthographic boundaries.

\section{Conclusion}

We introduced Serbian digraphia as a controlled evaluation paradigm for testing whether learned concept representations capture abstract semantics or remain tied to script-specific token patterns. Our experiments across the Gemma model family demonstrate that SAE-learned representations are substantially script-invariant: identical sentences across scripts activate overlapping feature sets far exceeding random baselines, with script variation introducing less representational divergence than paraphrasing. This property strengthens with model scale, as larger models exhibit more robust script-independent representations.

These findings suggest that SAE features can capture meaning at a level of abstraction that transcends surface tokenization, supporting their potential as interpretable, generalizable concept representations. We hope this work provides a useful foundation for further investigation into script-invariance and orthographic abstraction in neural networks.

\bibliography{references}
\bibliographystyle{iclr2026/iclr2026_conference}

\appendix

\section*{LLM Usage Disclosure}

In accordance with ICLR's Code of Ethics, we disclose the following use of large language models in this work. We used Claude (Anthropic) to assist with paper writing, including improving wording, grammar, and drafting sections. We also used LLMs to help generate experiment code.  All LLM-generated content, code, and results were thoroughly reviewed and verified by the authors, who take full responsibility for the accuracy and integrity of this submission.

\section{Dataset Details}
\label{app:dataset}

\subsection{Translation Methodology}

English sentences were translated to Serbian using the Google Translate V3 API, which returns translations in Cyrillic script by default. Serbian Latin versions were then generated via deterministic transliteration using Serbian's direct character correspondence. 

To ensure quality, we implemented several verification measures:
\begin{itemize}
    \item \textbf{Length validation:} All 90 English source sentences were constrained to 7--13 words to ensure comparable complexity.
    \item \textbf{Transliteration testing:} Round-trip tests (Latin $\rightarrow$ Cyrillic $\rightarrow$ Latin) verified mapping integrity against known test cases.
    \item \textbf{LLM verification:} Each translation batch was reviewed by Claude Sonnet 4 (using extended thinking mode) for accuracy, natural phrasing, and semantic drift. Additionally, predefined test cases were used to validate translation quality.
\end{itemize}

\subsection{Tokenization Analysis}

A potential confound in our analysis is that Latin and Cyrillic scripts might tokenize differently, which could artificially drive differences in SAE activations independent of semantic content. We addressed this concern through two analyses.

\begin{figure}[h]
\centering
\includegraphics[width=0.8\linewidth]{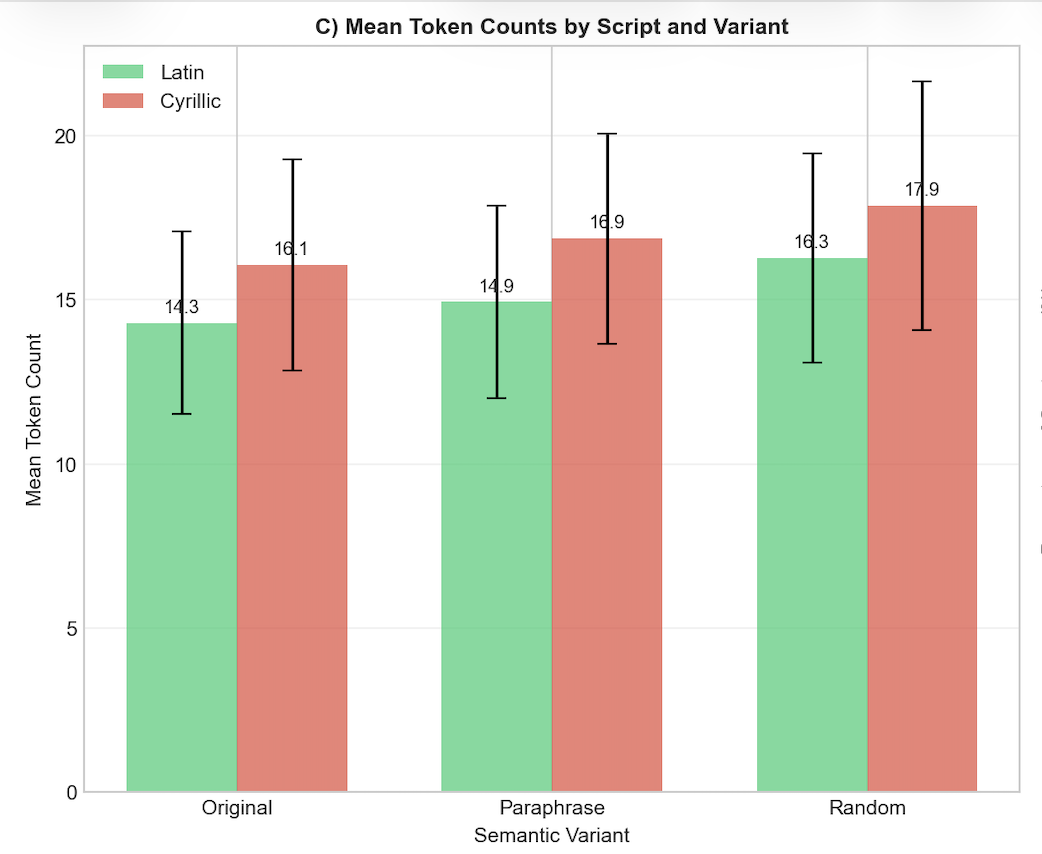}
\caption{Mean token counts by script and sentence type.}
\label{fig:token_counts}
\end{figure}

\paragraph{Token Count Comparison.}
We computed mean token counts for each script across all semantic variants (original, paraphrase, random). As shown in Figure~\ref{fig:token_counts}, token counts are nearly identical between scripts for the same content, with differences of only 1--2 tokens on average. This indicates that the Gemma tokenizer does not systematically produce longer or shorter sequences for one script over the other.

\begin{figure}[h]
\centering
\includegraphics[width=0.8\linewidth]{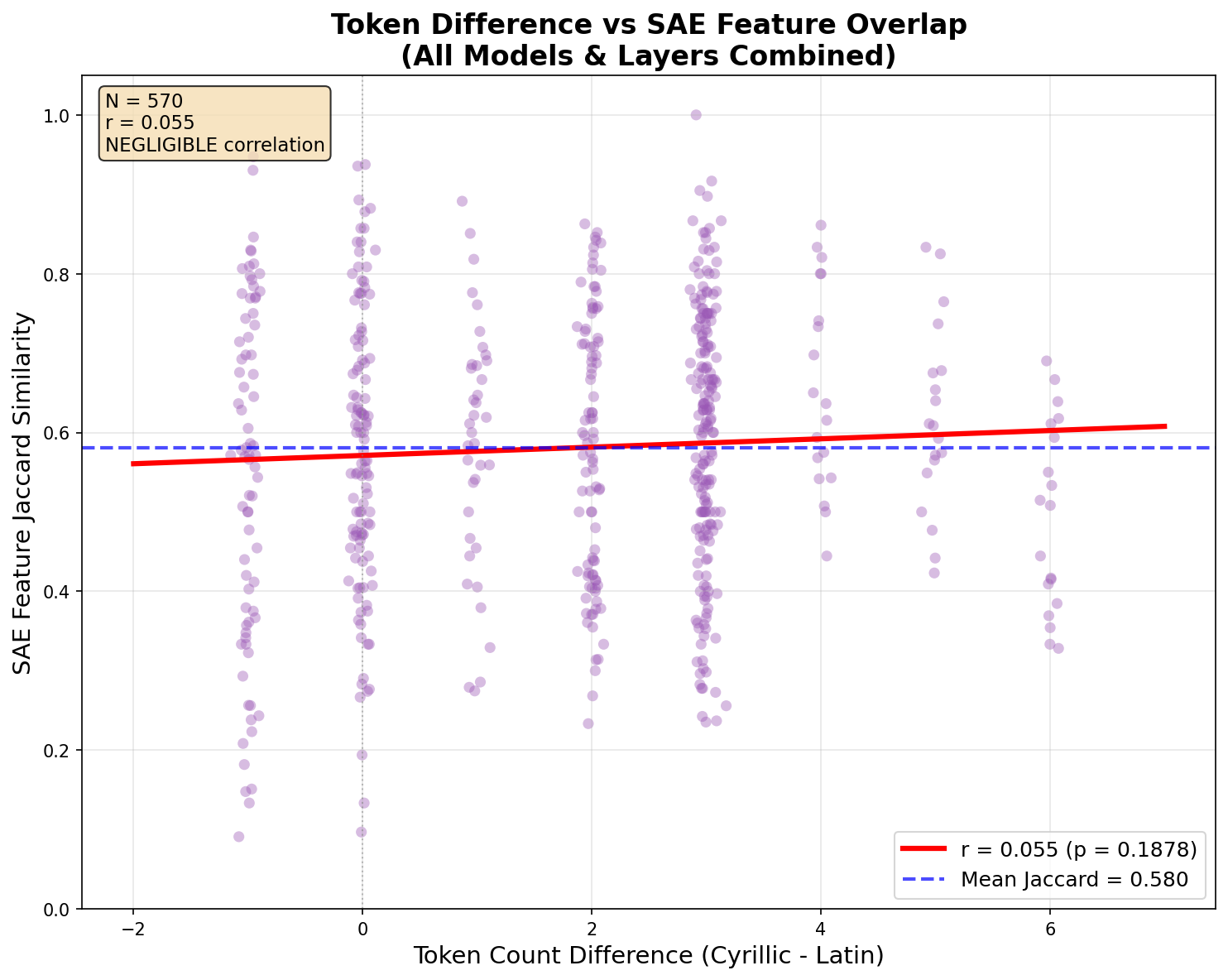}
\caption{Token count difference vs.\ SAE feature Jaccard similarity.}
\label{fig:token_vs_jaccard}
\end{figure}

\paragraph{Token Difference vs.\ Feature Similarity.}
To directly test whether tokenization differences predict feature overlap, we plotted the token count difference (Cyrillic $-$ Latin) against SAE feature Jaccard similarity for each sentence pair. As shown in Figure~\ref{fig:token_vs_jaccard}, there is no meaningful relationship ($r = 0.055$, $p = 0.188$). Sentences where Cyrillic uses more tokens than Latin do not show systematically different feature overlap compared to sentences with identical token counts. 

These results rule out tokenization as a confounding explanation for our cross-script similarity findings.

\subsection{Phrase Similarity Verification}

To independently verify that our sentence pairs exhibit the expected semantic relationships, we computed sentence embeddings using LaBSE (Language-agnostic BERT Sentence Embeddings). LaBSE is a multilingual model trained on 109 languages designed to map semantically equivalent sentences to similar embeddings regardless of language or script.

We analyzed three types of sentence similarities:
\begin{itemize}
    \item \textbf{Cross-script:} Serbian Latin $\leftrightarrow$ Serbian Cyrillic (same meaning, different scripts)
    \item \textbf{Cross-language:} English $\leftrightarrow$ Serbian Latin/Cyrillic (same meaning, different languages)
    \item \textbf{Paraphrase:} Original sentences vs.\ semantically equivalent paraphrases (same script and meaning, different wording)
\end{itemize}

As a control, we also computed random sentence similarity within scripts (semantically unrelated pairs) to establish a baseline.

\begin{figure}[h]
\centering
\includegraphics[width=\linewidth]{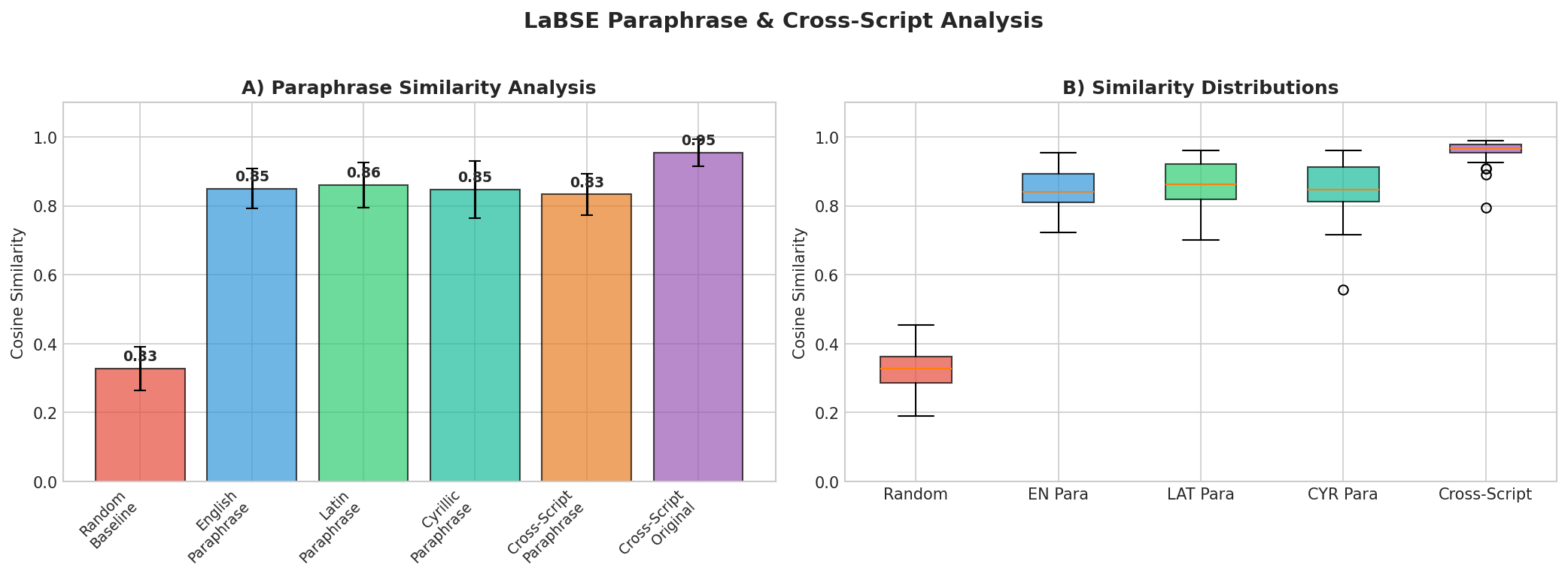}
\caption{LaBSE sentence similarity verification. (a) Histogram of similarity scores showing clear separation between random pairs and semantically related pairs. (b) Box plots of similarity distributions by condition. Cross-script original pairs achieve near-ceiling similarity, confirming semantic equivalence across scripts.}
\label{fig:labse}
\end{figure}

Figure~\ref{fig:labse} presents the results. Part (a) shows a histogram of similarity scores across conditions: random pairs cluster at low similarity, paraphrase pairs (English, Serbian Latin, Serbian Cyrillic) show high similarity, and cross-script original pairs achieve near-ceiling similarity. Part (b) displays the same data as box plots, revealing the distributions for each condition. Cross-script original pairs exhibit the highest and tightest distribution ($>$0.95), confirming that Latin and Cyrillic versions are recognized as semantically identical by an independent model. Paraphrase pairs show high but slightly more variable similarity, as expected given lexical differences. These results validate that our dataset exhibits the intended semantic structure.

\section{Experimental Results}
\label{app:results}
\subsection{Baseline Validation}
\label{app:baseline}
Baseline validation results averaged across all models and layers. Original vs.\ paraphrase similarity consistently exceeds original vs.\ random similarity across all conditions, confirming that SAE features capture semantic content.
\begin{center}
Table 2: Baseline validation results.\\[2mm]
\small
\begin{tabular}{lcc}
\toprule
\textbf{Condition} & \textbf{Orig $\leftrightarrow$ Para} & \textbf{Orig $\leftrightarrow$ Rand} \\
\midrule
English & 0.53 & 0.25 \\
Serbian Latin & 0.54 & 0.31 \\
Serbian Cyrillic & 0.54 & 0.31 \\
\bottomrule
\end{tabular}
\end{center}
\vspace{-2mm}
\subsection{Full Results by Model}
This appendix presents Jaccard similarity results averaged across all tested layers for each model. Abbreviations: EN = English, SR-Lat = Serbian Latin, SR-Cyr = Serbian Cyrillic.
\begin{center}
Table 3: Within-language comparisons.\\[2mm]
\footnotesize
\begin{tabular}{lccccc}
\toprule
\textbf{Comparison} & \textbf{270M} & \textbf{1B} & \textbf{4B} & \textbf{12B} & \textbf{27B} \\
\midrule
EN: Orig vs Para & 0.629 & 0.546 & 0.498 & 0.513 & 0.477 \\
EN: Orig vs Rand & 0.351 & 0.251 & 0.221 & 0.221 & 0.193 \\
SR-Lat: Orig vs Para & 0.595 & 0.555 & 0.526 & 0.526 & 0.496 \\
SR-Lat: Orig vs Rand & 0.493 & 0.341 & 0.262 & 0.235 & 0.210 \\
SR-Cyr: Orig vs Para & 0.634 & 0.537 & 0.520 & 0.516 & 0.510 \\
SR-Cyr: Orig vs Rand & 0.469 & 0.365 & 0.267 & 0.242 & 0.218 \\
\bottomrule
\end{tabular}
\end{center}
\vspace{-2mm}
\begin{center}
Table 4: Cross-script and cross-language comparisons.\\[2mm]
\footnotesize
\begin{tabular}{lccccc}
\toprule
\textbf{Comparison} & \textbf{270M} & \textbf{1B} & \textbf{4B} & \textbf{12B} & \textbf{27B} \\
\midrule
Cross-Script Orig & 0.501 & 0.537 & 0.571 & 0.624 & 0.649 \\
Cross-Script Para & 0.549 & 0.547 & 0.585 & 0.626 & 0.645 \\
\midrule
Lat Orig vs Cyr Para & 0.495 & 0.468 & 0.457 & 0.480 & 0.470 \\
Cyr Orig vs Lat Para & 0.488 & 0.475 & 0.461 & 0.475 & 0.468 \\
\midrule
Lat Orig vs Cyr Rand & 0.421 & 0.324 & 0.253 & 0.233 & 0.211 \\
Cyr Orig vs Lat Rand & 0.413 & 0.317 & 0.239 & 0.225 & 0.210 \\
\midrule
Lat Orig vs EN Rand & 0.251 & 0.199 & 0.180 & 0.173 & 0.164 \\
Cyr Orig vs EN Rand & 0.248 & 0.196 & 0.162 & 0.161 & 0.159 \\
\bottomrule
\end{tabular}
\end{center}

\end{document}